\documentclass{esannV2}
\usepackage[pdftex]{graphicx}
\usepackage[latin1]{inputenc}
\usepackage{booktabs}
\usepackage{microtype}
\usepackage{subcaption}
\usepackage{url}
\usepackage{amssymb,amsmath,array}
\usepackage{hyperref}
\begin{document}

\title{A Protocol for Continual Explanation of SHAP}

\author{Andrea Cossu$^{1, 2}$, Francesco Spinnato$^{2, 3}$, Riccardo Guidotti$^1$ and Davide Bacciu$^1$
%
\thanks{Work supported by: EU EIC project EMERGE (Grant No. 101070918), EU NextGenerationEU programme under the funding schemes PNRR-PE-AI FAIR (Future Artificial Intelligence Research), PNRR-SoBigData.it - Strengthening the Italian RI for Social Mining and Big Data Analytics - Prot. IR0000013, H2020-INFRAIA-2019-1: Res. Infr. G.A. 871042 \emph{SoBigData++}, G.A. 761758 \emph{Humane AI}, G.A. 952215 \emph{TAILOR}, ERC-2018-ADG G.A. 834756 \textit{XAI}, and CHIST-ERA-19-XAI-010 SAI, and by the Green.Dat.AI Horizon Europe research and innovation programme, G.A. 101070416.}
%
\vspace{.3cm}\\
1- University of Pisa;  2- Scuola Normale Superiore; 3- CNR. Pisa, Italy
}
%


\maketitle

\begin{abstract} 
Continual Learning trains models on a stream of data, with the aim of learning new information without forgetting previous knowledge. Given the dynamic nature of such environments, explaining the predictions of these models can be challenging. We study the behavior of SHAP values explanations in Continual Learning and propose an evaluation protocol to robustly assess the change of explanations in Class-Incremental scenarios. We observed that, while Replay strategies enforce the stability of SHAP values in feedforward/convolutional models, they are not able to do the same with fully-trained recurrent models. We show that alternative recurrent approaches, like randomized recurrent models, are more effective in keeping the explanations stable over time.
\end{abstract}

\section{Introduction}
Continual Learning (CL) studies the challenges of training models in dynamic environments, where data distribution drifts over time. When trained on such non-stationary data, neural networks have been shown to \emph{forget} previous knowledge. Explaining neural network predictions is an inherently difficult task at the center of eXplainable Artificial Intelligence (XAI)~\cite{guidotti2018survey}, and can be even harder when models are updated on drifting distributions. 
How will the explanations change in a CL scenario? How the effect of non-stationarity can be shown in an explanation?


There are very few works at the intersection of CL and XAI. In \cite{ebrahimi2021}, the authors mitigate forgetting by enforcing visual explanations (i.e., feature maps) stability during CL with Replay. More recently, \cite{rymarczyk2023} introduced a regularization-based CL approach to keep prototypes stable as new classes are learned. In both works, the focus is on the design of a CL approach that mitigates forgetting. Here, we are interested in quantifying the degree of change of an explanation and studying the role played by the data domain (images vs. sequences), the model architecture (feedforward/convolutional vs. recurrent), and the type of CL strategy.

In this paper, we tackle XAI for CL 
by quantitatively and qualitatively assessing the behavior of one of the most used XAI techniques, i.e., SHAP (SHapley Additive exPlanations)~\cite{lundberg2017unified}, in a supervised CL scenario (Class-Incremental) where new classes are encountered over time. 
Further, we propose an evaluation protocol (including two metrics) that takes into account the non-stationary nature of the environment. 
It considers a candidate explanation \emph{for each} possible class and makes a comparison across all of them. In this way, we are not only looking at the explanation associated with the \emph{target} class. In fact, in a CL scenario, there is no guarantee that the model will always be able to classify the same example correctly. Moreover, at test/deployment time, the correct class is unknown. Our protocol breaks free from trusting the model prediction or from relying on external information. 

We assessed our evaluation protocol on $3$ CL benchmarks, from the computer vision and audio processing domains. We compare the drift of explanations when using CL strategies with respect to the model jointly trained on all data at once. We discovered that Replay strategies are effective in preventing drift, as expected. However, the behavior of Recurrent Neural Networks (RNNs) is peculiar, since, surprisingly, explanations drift \emph{more} with Replay than with no CL strategy. We show that the issue is related to the \emph{fully-trained} recurrent component. In fact, we were able to prevent explanation drift by using a random recurrent model, without losing predictive accuracy. To the best of our knowledge, we are the first to study how SHAP values change in a CL setting and to highlight the peculiar case represented by explanations in RNNs.

\section{Continual Explanation}
In a CL scenario, a model $h$ faces a stream of experiences $e_1, e_2, \ldots$, where each $e_i$ corresponds to a dataset of input-target pairs $D_i = (x_j, y_j)_{j=1}^{N_e}$ (supervised CL). The model is trained on one experience after another. We focus on the popular Class-Incremental scenario, where each experience introduces a new set of target classes that never reoccur in future experiences. This scenario is well-known to induce a large amount of forgetting in the model: after training on an experience $e_i$, the accuracy on the test set of experiences $e_j, j<i$ largely deteriorates.

Our proposed evaluation protocol uses SHAP to explain the importance of each input feature, i.e., the contribution of each feature toward the model prediction. In simple terms, SHAP computes this contribution by perturbing the input instance $x$ using the mask $z' \in \{0,1\}^{K}$ to decide which of the $K$ feature values to keep or replace in $x$. The contribution of each feature to the model output is computed by observing how the model output changes depending on different perturbations. In this way, a positive or negative SHAP value $\phi$ is assigned to each feature value, and the resulting explanation is represented as an additive feature attribution method. Formally:
$g(z') = \phi_0 + \sum_{k=1}^{K} \phi_k z'_k$.
\noindent In other words, given an input $x$, the explanation model $g$ tries to linearly approximate the output of a model $f$ in the local neighborhood obtained by perturbing $x$, i.e., $g(z') \approx f(x)$. The term $\phi_0$ denotes the base value, i.e., the default prediction for an ``empty'' instance. Since we compute SHAP against each possible class, we only consider positive SHAP values, by clamping negative values to 0. This allows us to investigate how much each input feature (e.g., pixels in images) drives the prediction toward the candidate class. In such a scenario, we compute the SHAP values at the end of training on each experience. 

As a ``ground truth'' reference, we compute the SHAP values of the model jointly trained on the union of all the experiences, in an offline fashion (Joint Training). This model is immune to forgetting since it does not learn continuously.
To evaluate the performance of different CL strategies, we propose two novel metrics. The first metric $M$ measures how much the SHAP values drift from the ones of the jointly trained model.
For each candidate class (output unit of the model), $M$ is defined as follows:
\begin{equation} \label{eq:metric-sum}
    M(S, J) = \frac{1}{K} \bigg(\sum_{i=1}^K S_i - \sum_{i=1}^K J_i\bigg)^2,
\end{equation}
where $S$ contains the positive SHAP values for a given strategy S and $J$ contains the positive SHAP values for the Joint Training strategy. Both $S$ and $J$ contains $K$ elements. 

For computer vision experiments, the metric $M$ of Eq. \ref{eq:metric-sum} may fail to capture the spatial dependencies between pixels. The sum, in fact, compares SHAP values independently of their position in an image. We monitor spatial locality by proposing a second metric:
\begin{equation} \label{eq:metric-pool}
    M_{\text{pool}}(S, J) = \frac{1}{P} \sum_{i=1}^P (\text{pool}(S, 4)_i - \text{pool}(J, 4)_i)^2,
\end{equation}
where ``pool'' represents a 2D average pooling operation with a kernel of 4x4 pixels and $P$ is the number of pixels produced by the pooling operation. SHAP values are normalized to have zero mean and unitary variance. The normalization allows us to compare the distribution of the positive values instead of the scale (which is taken into account by the metric $M$ in Eq. \ref{eq:metric-sum}). Metric $M_{\text{pool}}$ includes spatial locality since it compares neighborhoods of 4x4 pixels from one strategy against the corresponding neighbor from Joint Training. 

\section{Experiments}
For our experiments, we choose two CL benchmarks from computer vision: Split MNIST and Split CIFAR-10. Both of them have 5 experiences, with 2 classes in each experience. In the same setup, we also study the Synthethic Speech Commands (SSC) benchmark \cite{cossu2021b}, which is composed of pre-processed audio sequences of $101$ steps with $40$ Mel features. We leverage a feedforward network on MNIST, a ResNet18 on CIFAR-10, and an LSTM on SSC.

We employ two Replay strategies, GSS \cite{aljundi2019a} and Experience Replay (ER), that mitigate forgetting by augmenting each mini-batch from the current experience with a sample from a fixed-size memory of examples from previous experiences. Also, we use the Naive strategy, a lower bound baseline that simply fine-tunes the model $h$ on the stream of experiences without any CL technique. We ensure that CL models achieve a performance (average accuracy on the test sets of all experiences after training on the last experience) that is on par with the expected CL performance for each strategy.

We compute SHAP values for the examples in the \emph{first experience} $e_1$, since it is the experience that suffers forgetting the most after training on the entire stream. We use $600$ examples as background and $50$ per class as a test, taken from the test set of $e_1$. For each strategy (Naive, GSS, ER), we compare the SHAP values w.r.t. the model jointly trained on $\cup_i D_i$. For our implementation, we use the Avalanche library \cite{carta2023} and GradientSHAP~\cite{lundberg2017unified}, i.e., a SHAP version tailored to neural networks\footnote{The code to reproduce all experiments can be found at \href{https://github.com/AndreaCossu/explainable-continual-learning}{explainable-continual-learning}}. \\

\begin{figure}[t]
    \centering
    
    \begin{subfigure}{0.85\textwidth}
    \includegraphics[width=0.95\textwidth]{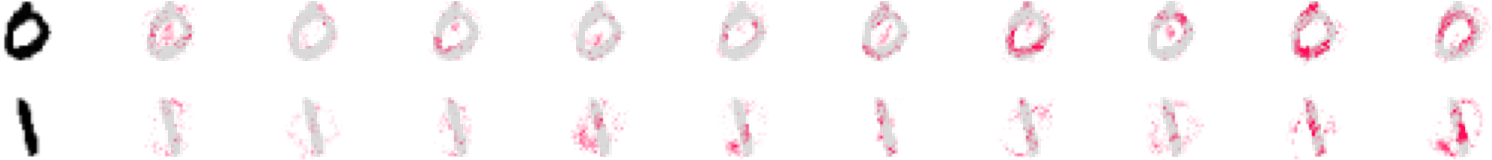}
    \caption{Naive}
    \end{subfigure}
    \begin{subfigure}{0.85\textwidth}
    \includegraphics[width=0.95\textwidth]{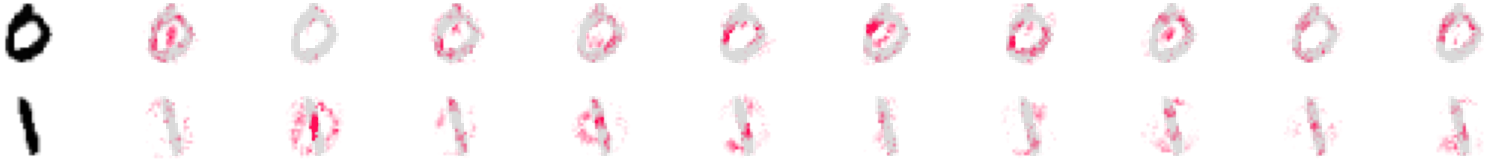}
    \caption{ER}
    \end{subfigure}

    \caption{Positive SHAP values after training on the last experience. The first column shows the input image, and the other ten columns show the SHAP values for each class (the more active, the higher the pixel's contribution toward the class). Replay is able to effectively preserve the SHAP values for class 0 and class 1, while Naive is mostly activated by class 8 and 9 (forgetting).}
    \label{fig:mnist}
\end{figure}

Figure \ref{fig:mnist} shows the SHAP values saliency maps for each candidate class for MNIST. We can clearly see that the last columns (output units) of the images are more active in Naive than in ER. This means that the model is erroneously choosing the last two output units in place of the correct ones (the first two units). We observed the same behavior also on CIFAR-10. This, perhaps intuitive, result highlighted the role of forgetting in SHAP and motivated us to quantitatively measure its change with $M$ and $M_{\text{pool}}$.

\begin{figure}[!ht]
    \centering

    \begin{subfigure}{0.3\textwidth}
    \includegraphics[width=0.95\textwidth]{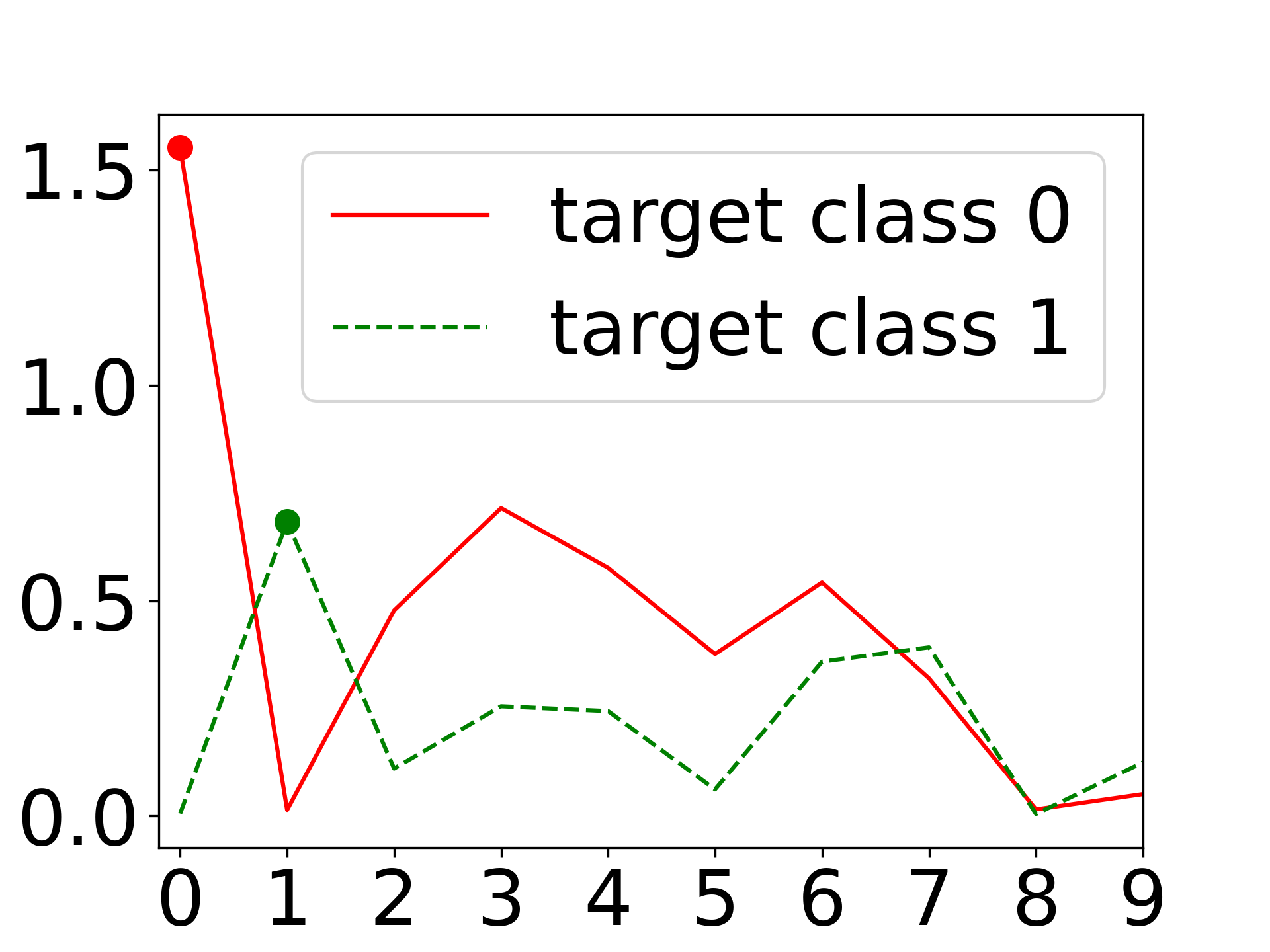}
    \caption{Naive on MNIST}
    \end{subfigure}
    \begin{subfigure}{0.3\textwidth}
    \includegraphics[width=0.95\textwidth]{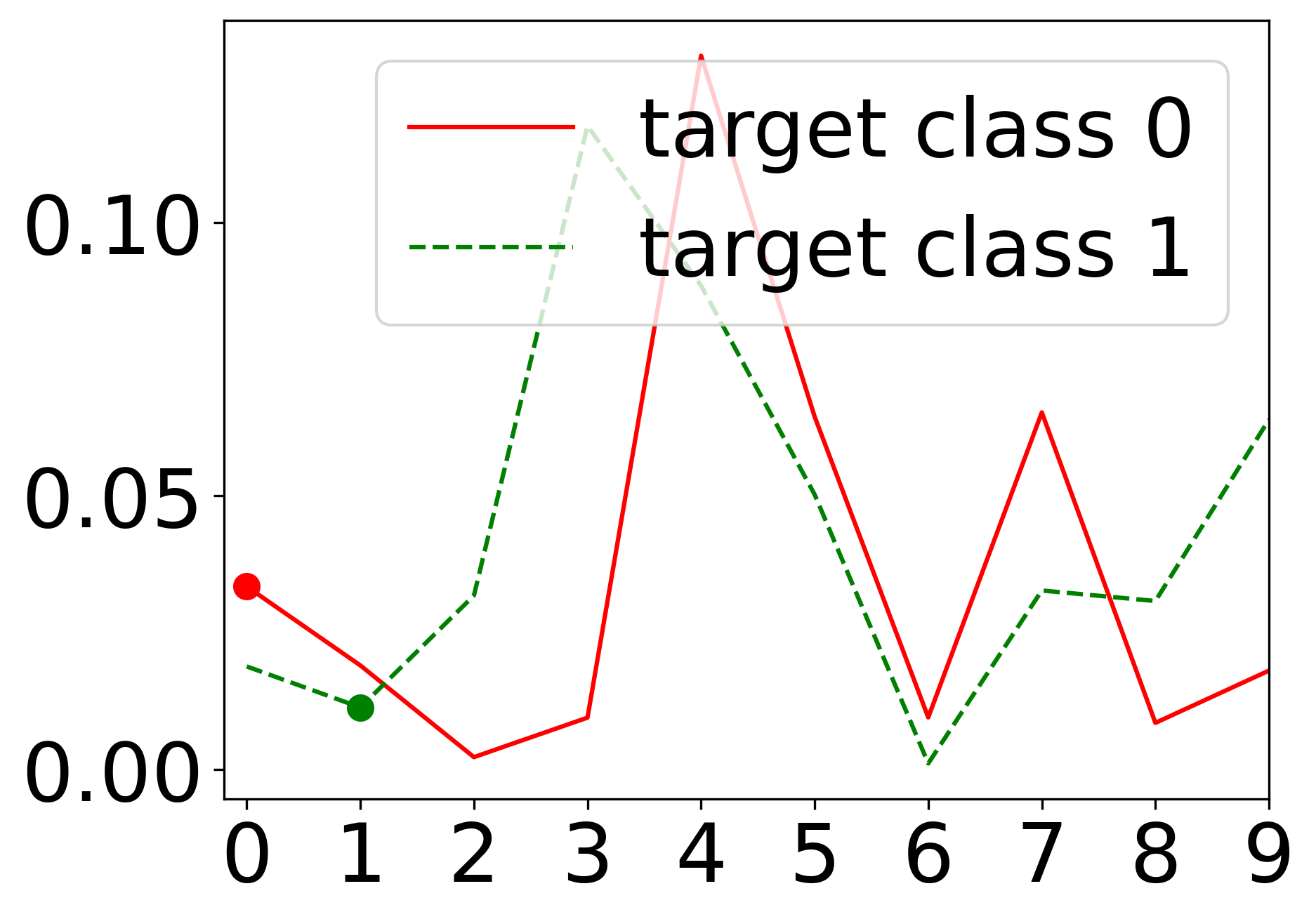}
    \caption{GSS on MNIST}
    \end{subfigure}
    \begin{subfigure}{0.3\textwidth}
    \includegraphics[width=0.95\textwidth]{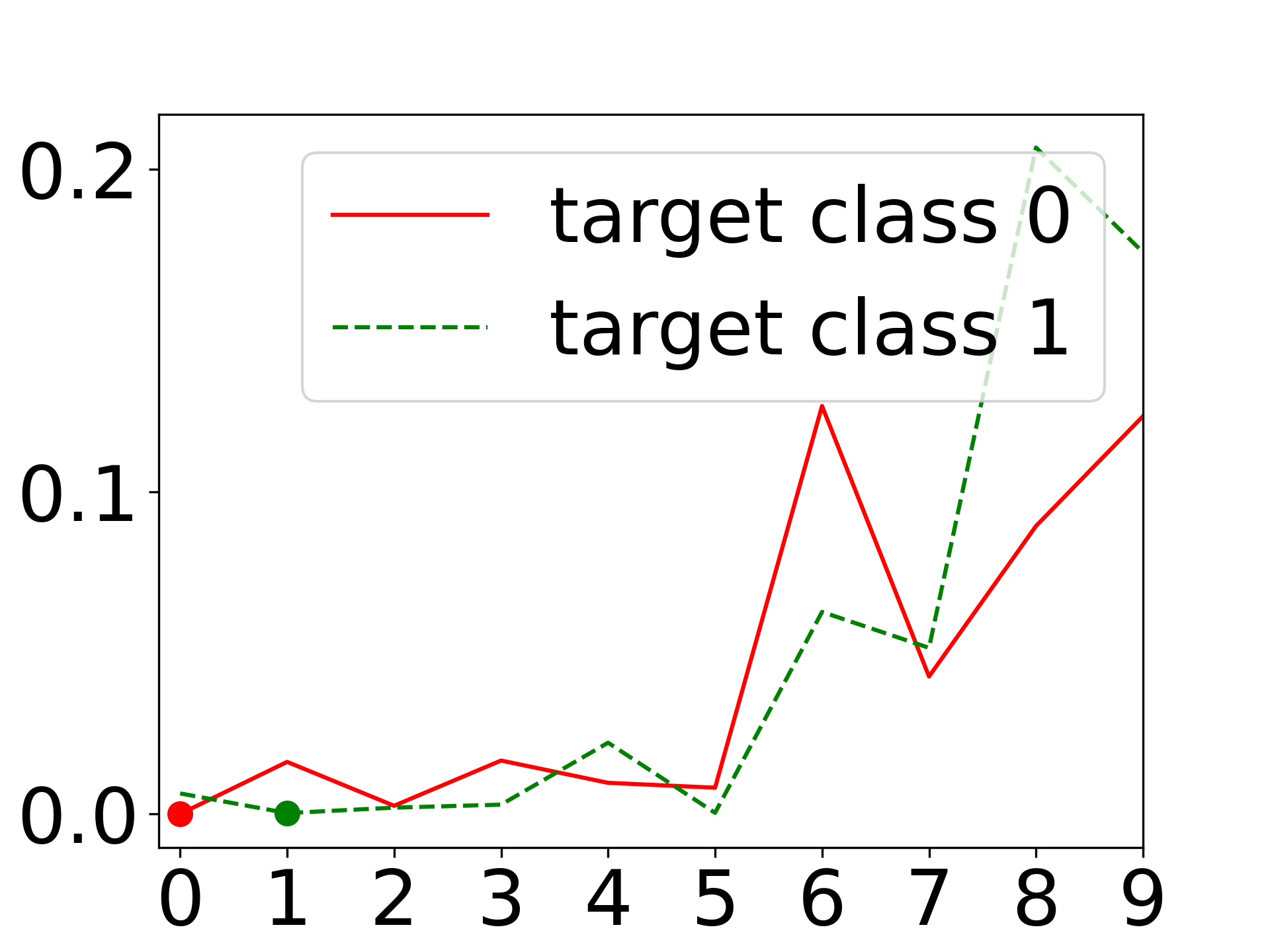}
    \caption{ER on MNIST}
    \end{subfigure}
    
    \begin{subfigure}{0.3\textwidth}
    \includegraphics[width=0.95\textwidth]{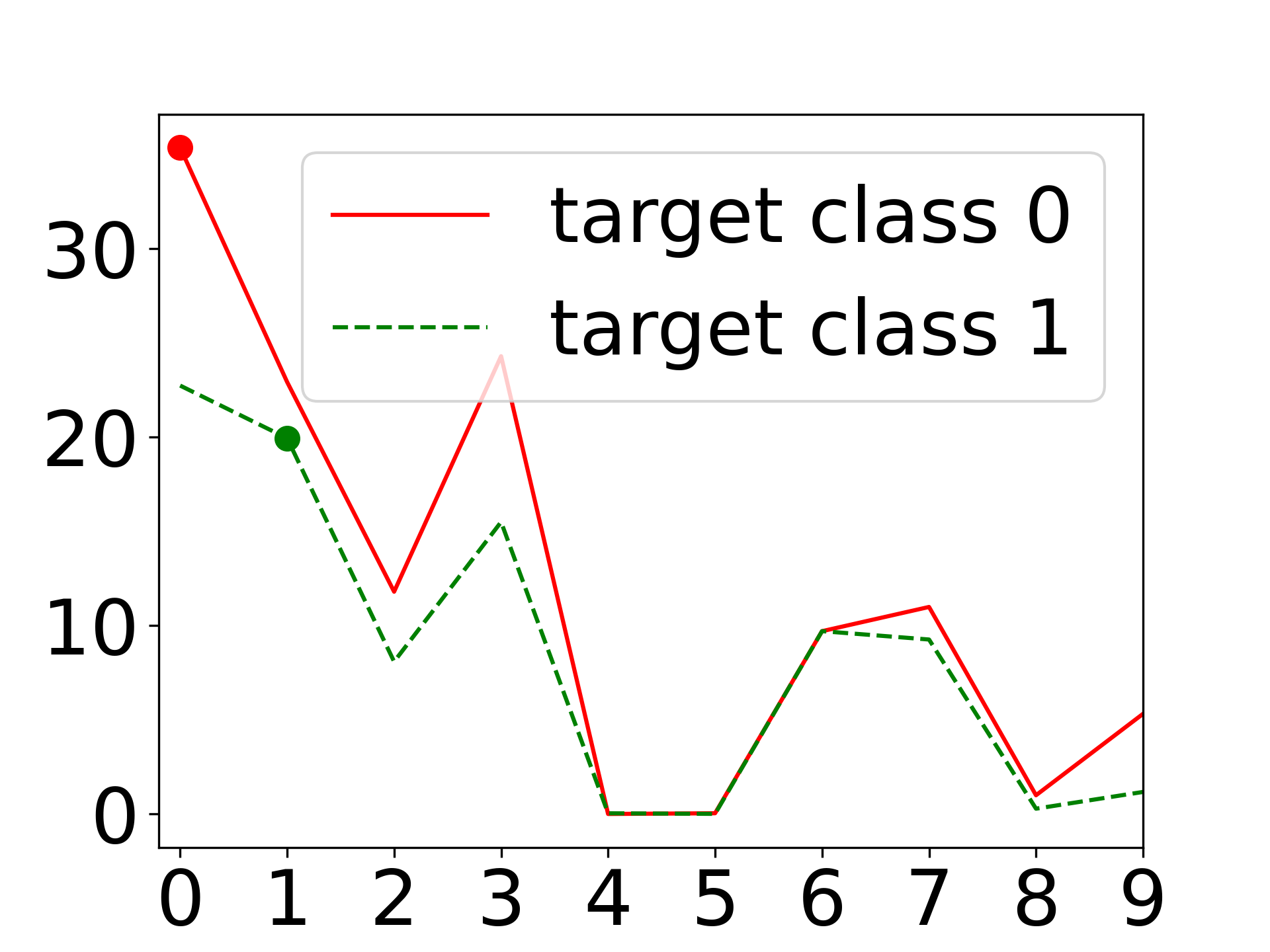}
    \caption{Naive on CIFAR}
    \end{subfigure}
    \begin{subfigure}{0.3\textwidth}
    \includegraphics[width=0.95\textwidth]{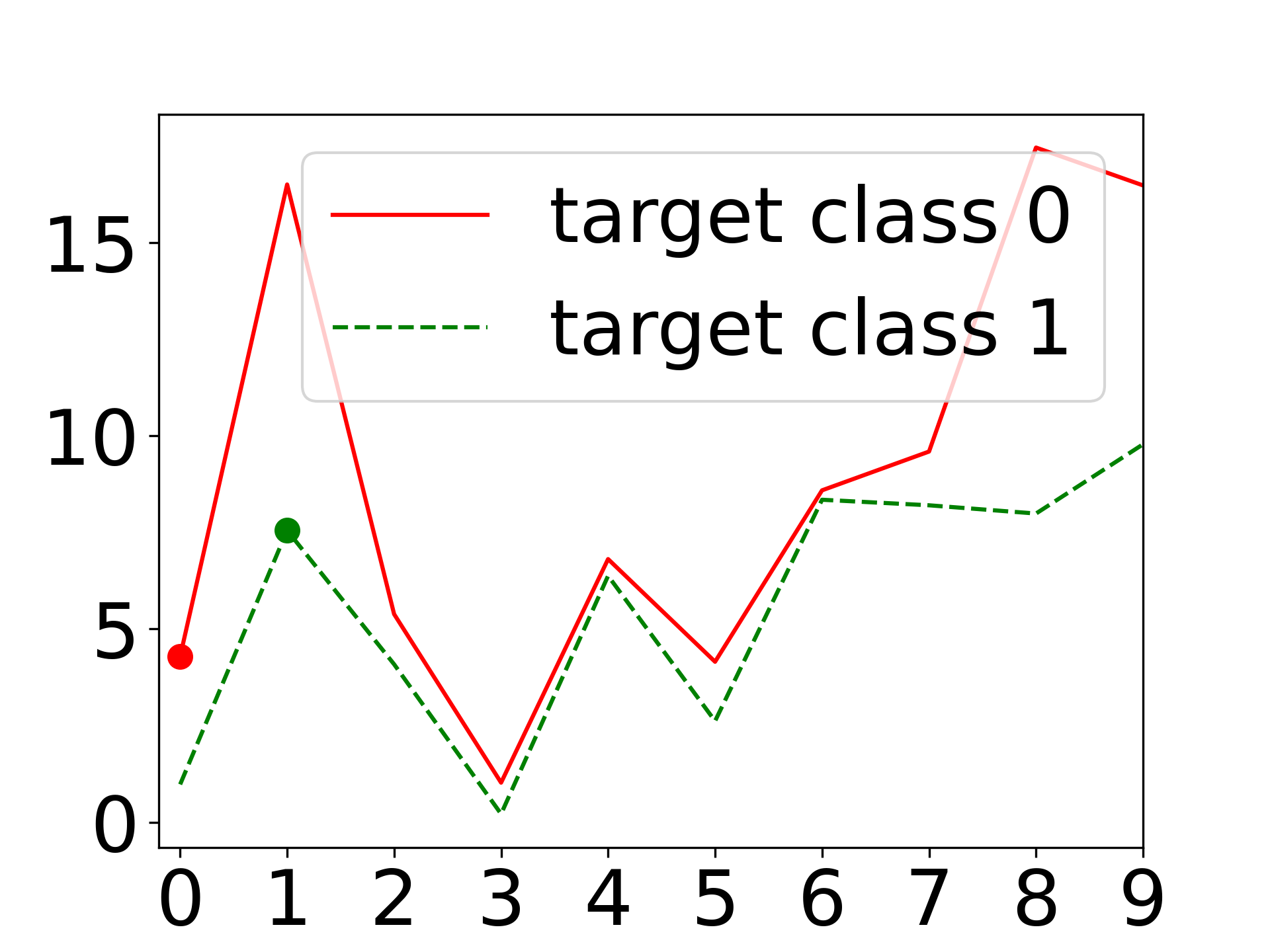}
    \caption{GSS on CIFAR}
    \end{subfigure}
    \begin{subfigure}{0.3\textwidth}
    \includegraphics[width=0.95\textwidth]{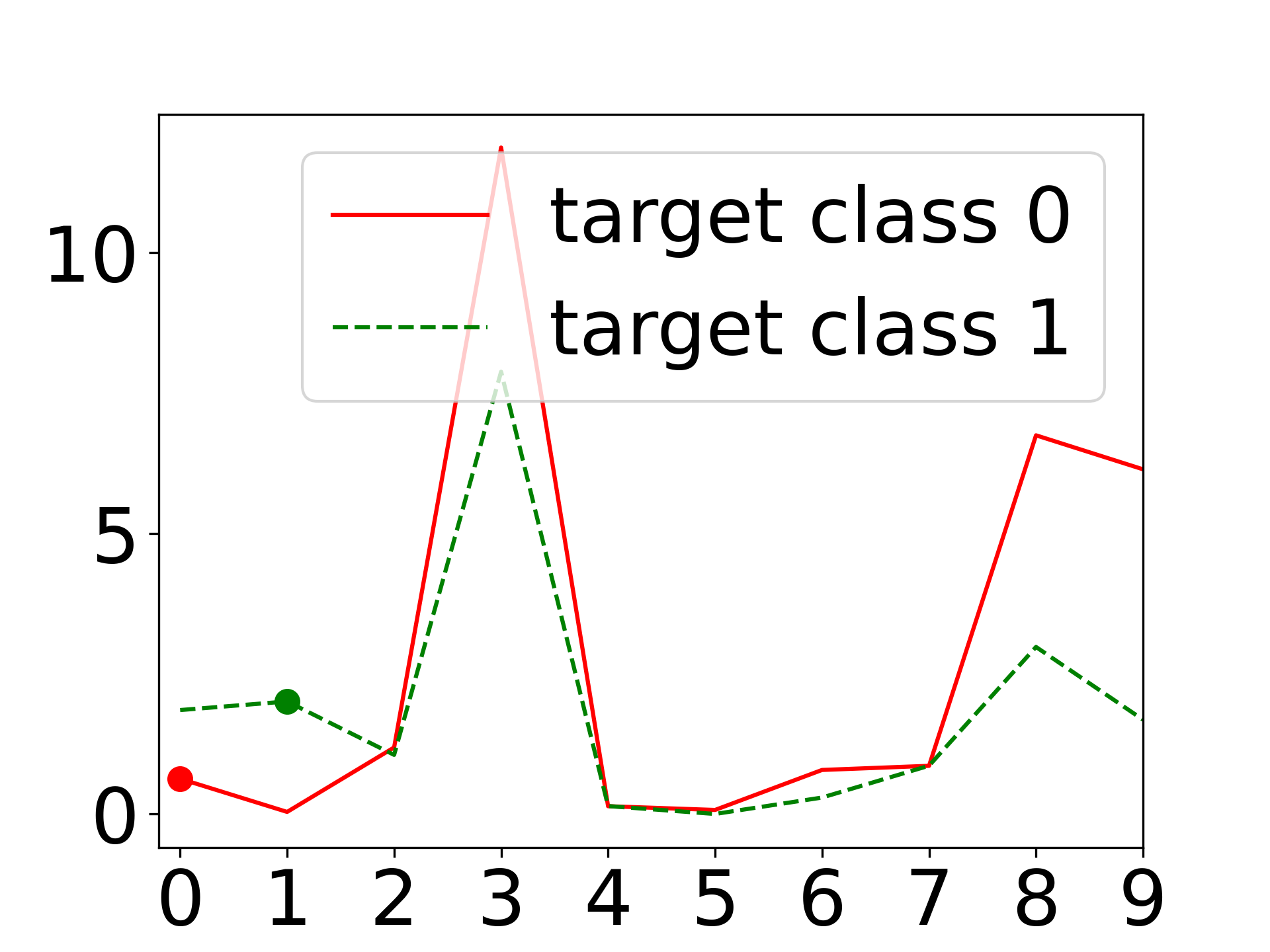}
    \caption{ER on CIFAR}
    \end{subfigure}
    
    \begin{subfigure}{0.3\textwidth}
    \includegraphics[width=0.95\textwidth]{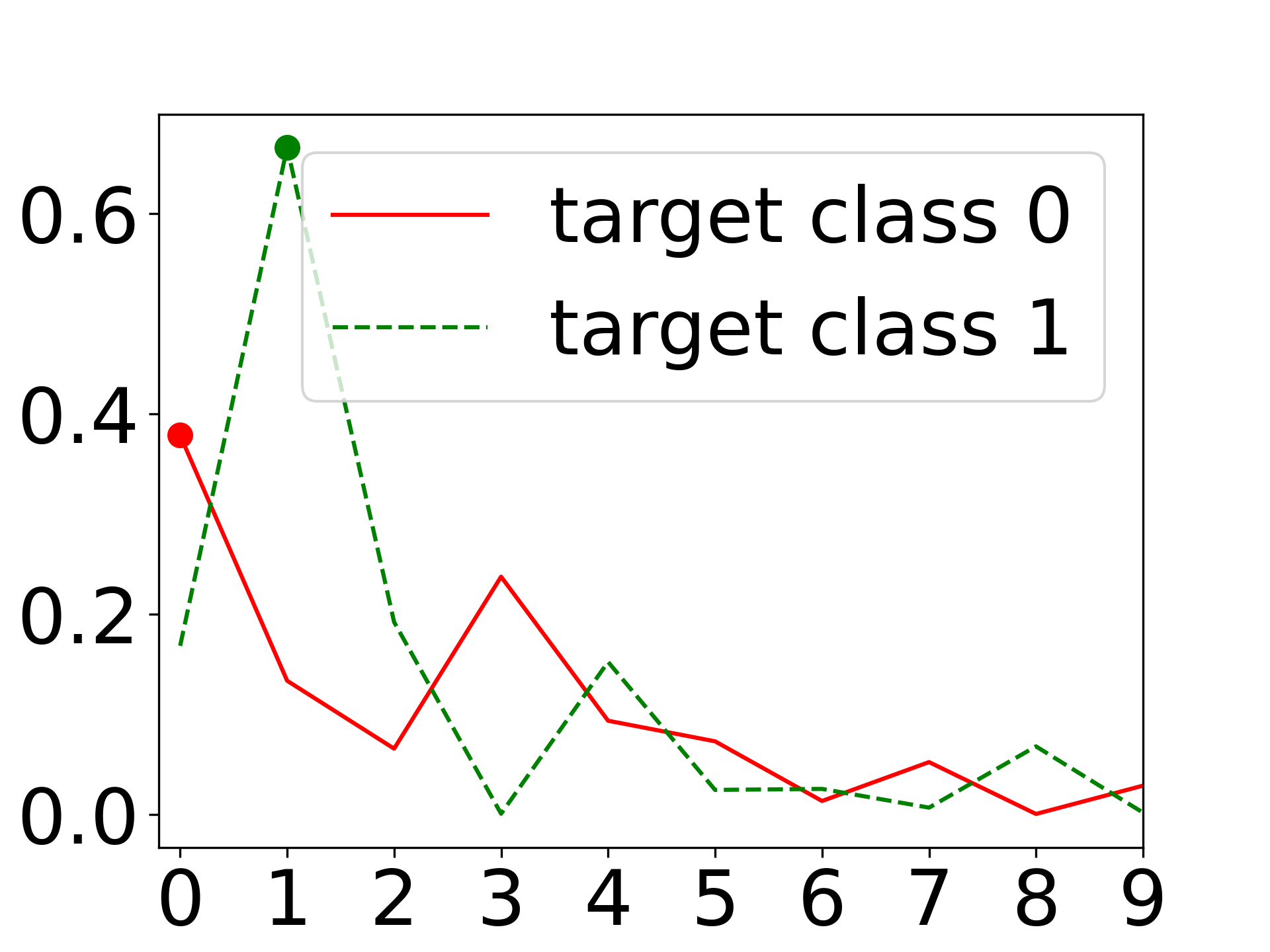}
    \caption{Naive on SSC - LSTM}
    \end{subfigure}
    \begin{subfigure}{0.3\textwidth}
    \includegraphics[width=0.95\textwidth]{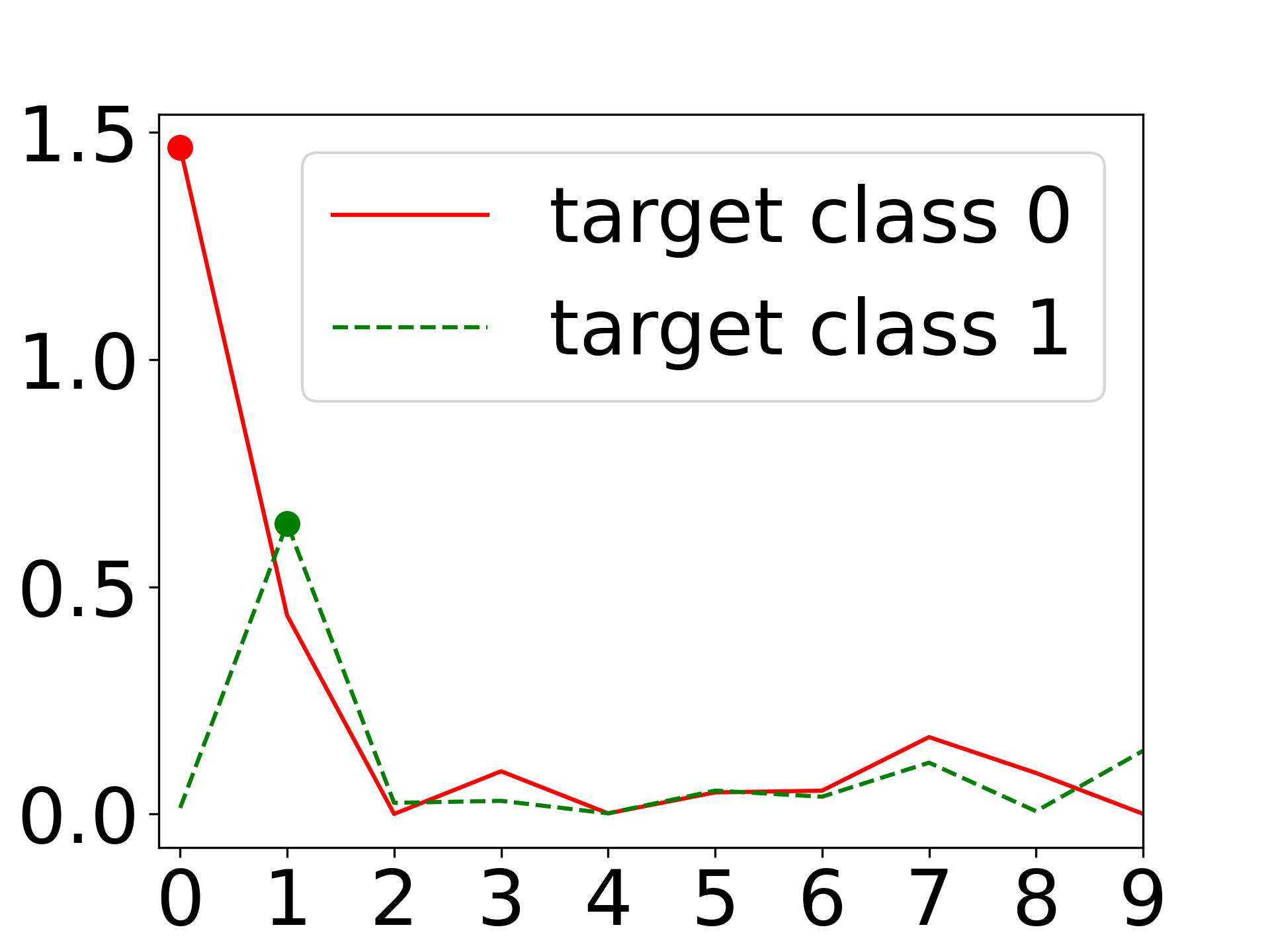}
    \caption{GSS on SSC - LSTM}
    \end{subfigure}
    \begin{subfigure}{0.3\textwidth}
    \includegraphics[width=0.95\textwidth]{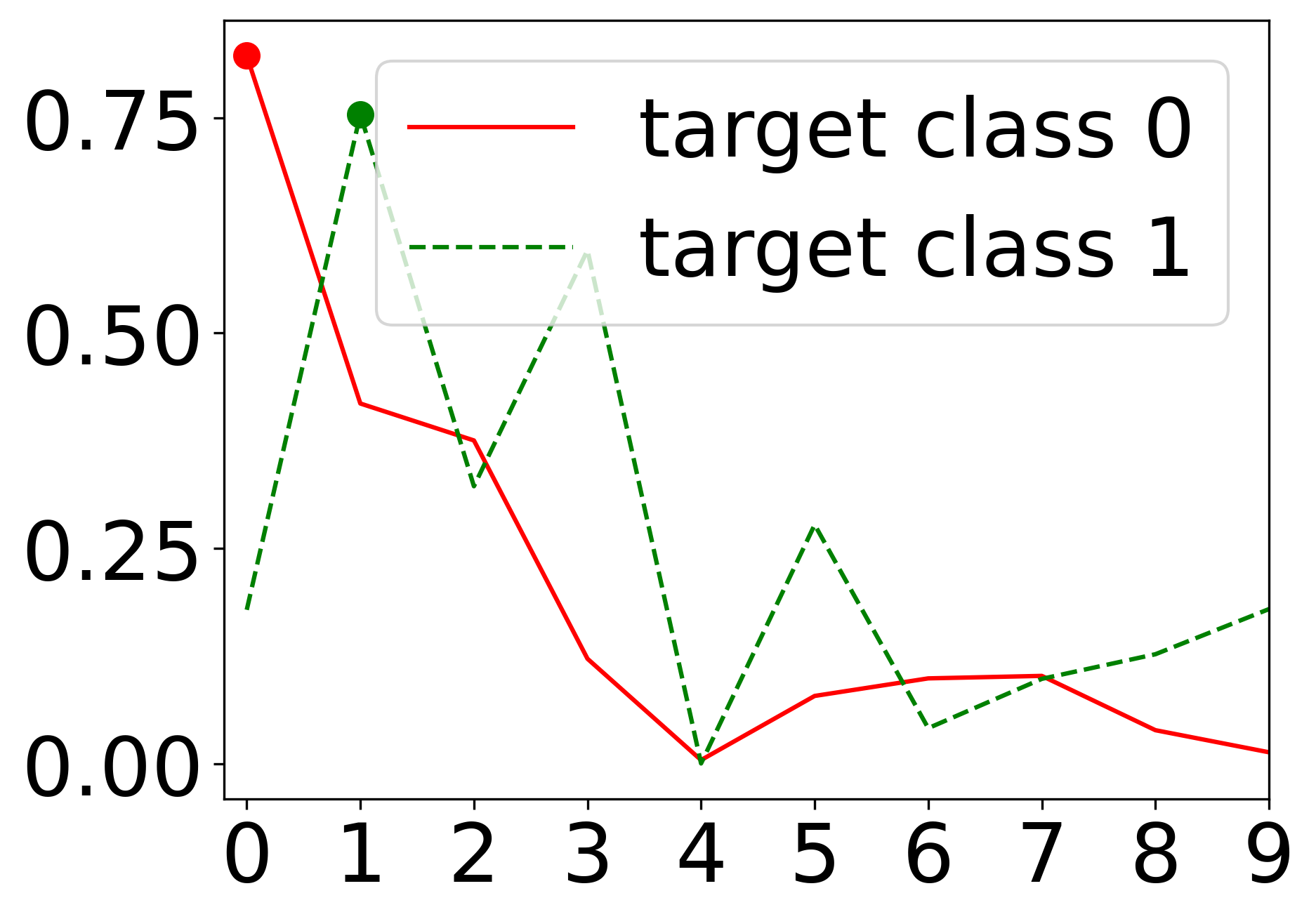}
    \caption{ER on SSC - LSTM}
    \end{subfigure}

    \begin{subfigure}{0.3\textwidth}
    \includegraphics[width=0.95\textwidth]{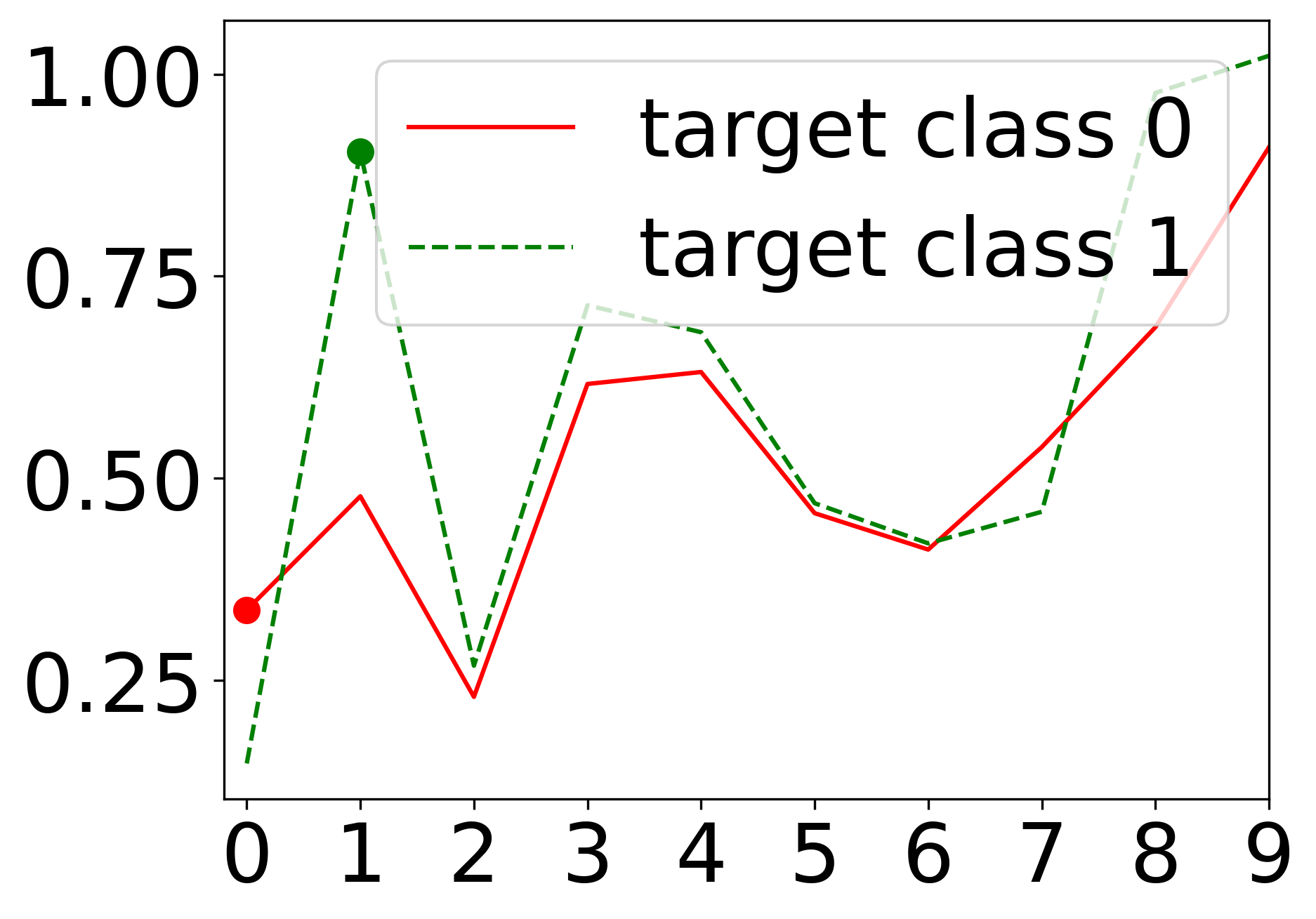}
    \caption{Naive on SSC - ESN}
    \end{subfigure}
    \begin{subfigure}{0.3\textwidth}
    \includegraphics[width=0.95\textwidth]{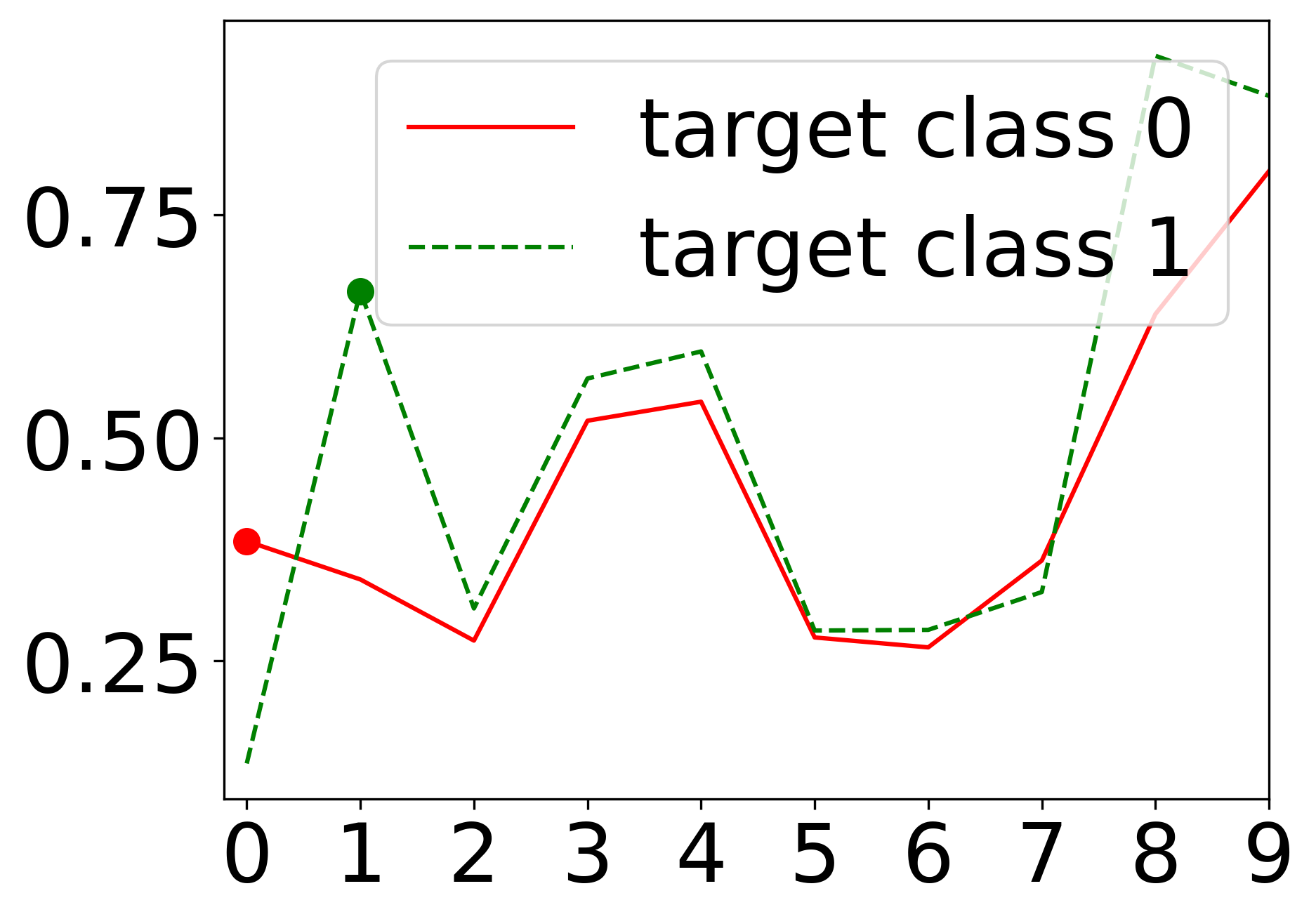}
    \caption{GSS on SSC - ESN}
    \end{subfigure}
    \begin{subfigure}{0.3\textwidth}
    \includegraphics[width=0.95\textwidth]{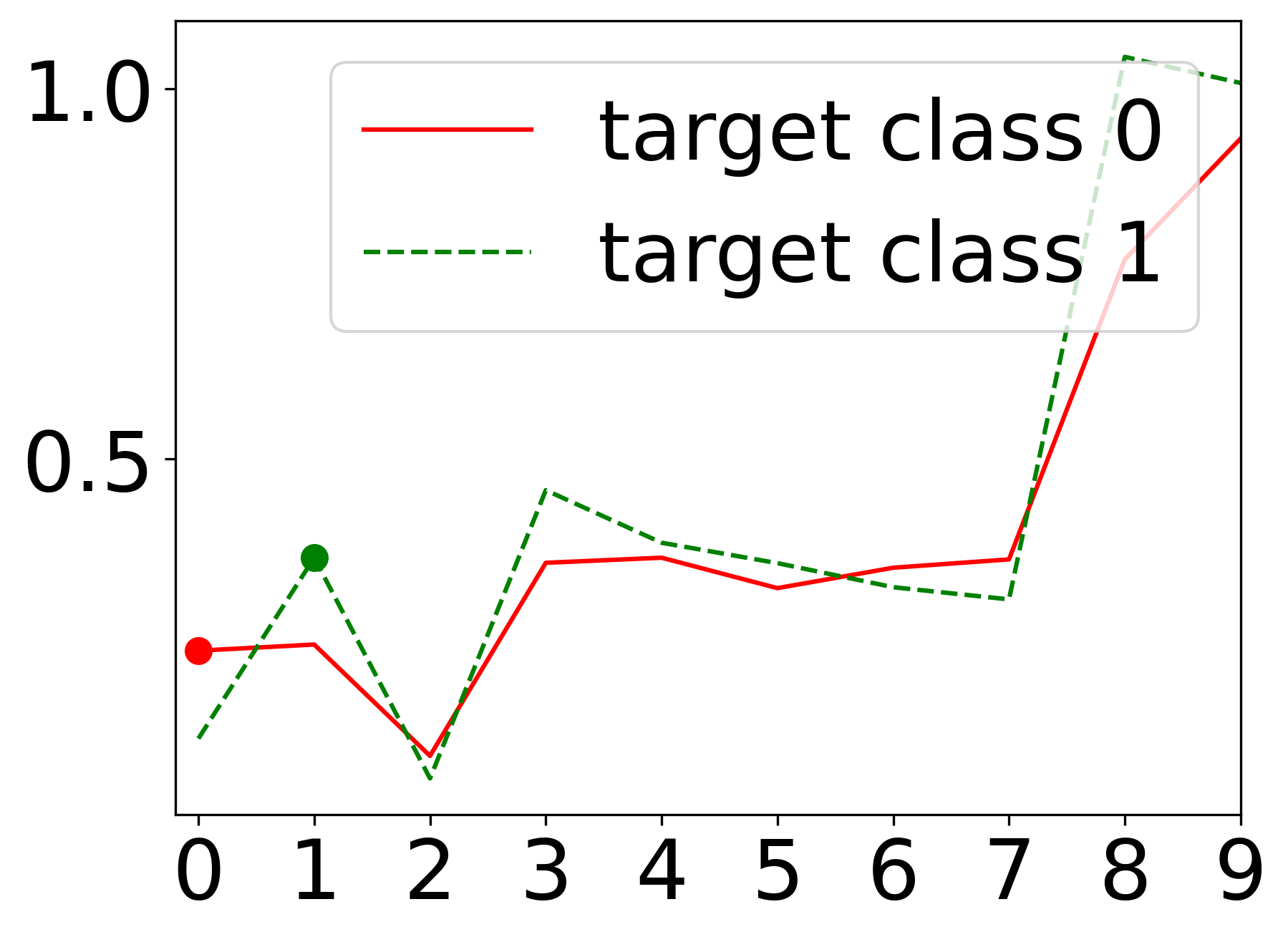}
    \caption{ER on SSC - ESN}
    \end{subfigure}

    \caption{Metric $M$ for each strategy and benchmark (lower is better). In each plot, the x-axis represents the number of classes in the dataset. The dots indicate the value of $M$ for the output unit corresponding to the target class. Each curve is averaged over $50$ examples from the test set.}
    \label{fig:mplots}
\end{figure}

Figure \ref{fig:mplots} shows results for metric $M$. The Naive strategy drifts much more than the Replay strategy for the target class (either 0 or 1, denoted by the dots). However, Naive usually better preserves SHAP values for the last classes (e.g., 8 and 9), where $M$ is closer to zero than in Replay. This is reasonable, since Naive blindly optimizes on the current experience up to the last one, while Replay strategies are constrained by the limited replay buffer containing examples from previous classes. Note that this is also aligned with the previously discussed results from Figure \ref{fig:mnist}. 
Similar results hold also for $M_{\text{pool}}$ in MNIST and CIFAR-10 (Table \ref{tab:mpool}), showing that Replay better preserves explanations even when considering the spatial locality of the pixels.

Interestingly, RNNs on SSC show a completely different behavior, as shown by the third row of Figure \ref{fig:mplots}. Although the Replay performance in terms of accuracy is \emph{better} than the one of Naive (80/90\% vs. 20\%), SHAP values of Replay strategies drift much more for RNNs. This counter intuitive result prove once more the unexpected behavior of fully-trained recurrent models in CL \cite{cossu2021b}. Inspired by \cite{cossu2021}, we ran experiments on SSC with Echo-State Networks (ESNs). These are RNNs with a fixed recurrent component. ESNs achieve comparable performance in terms of accuracy with respect to LSTM, while also preserving the stability of SHAP values (Figure \ref{fig:mplots}, last row). Similar results are obtained when using a shallow 1D Convolutional network in place of an RNN, pointing at the fact that the main issue in the drift of explanations comes from the \emph{trained} recurrent component in the LSTM and not from the recurrence itself.

\begin{table}[t]
    \centering
    \small
        \begin{tabular}{ccc|cc|cc}
        \toprule
                 & \multicolumn{2}{c}{Naive} & \multicolumn{2}{c}{GSS} & \multicolumn{2}{c}{ER} \\ \midrule
                 & C0          & C1          & C0         & C1         & C0         & C1        \\ \midrule
        MNIST    & 2.9e-4      & 2.5e-4      & 1.5e-4     & \textbf{1.4e-4}     & \textbf{1.4e-4}     & 1.5e-4    \\ \midrule
        CIFAR-10 & 1.1e-3      & 6.0e-4      & \textbf{1.6e-4}     & \textbf{1.5e-4}     & 2.8e-4     & 2.8e-4   \\ \bottomrule
        \end{tabular}
    \caption{Metric $M_{\text{pool}}$ for each strategy and vision benchmark (best $M_{\text{pool}}$ in bold for each class). The metric is shown for target classes 0 and 1 only (the dots in Figure \ref{fig:mplots}). The behavior on the other classes follows the one of Figure \ref{fig:mplots}.}
    \label{tab:mpool}
\end{table}

    
    
    

\section{Conclusion}
We studied how SHAP values change in a CL environment. Our evaluation protocol compares all candidate explanations (one for each candidate class) and introduces two metrics to measure the drift with respect to a jointly trained model. Replay strategies are effective in preserving the stability of explanations. However, when using fully-trained recurrent models (e.g., LSTMs), explanations drift from the ones of Joint Training, while prediction accuracy remains high. Similar behavior has been documented when looking only at the predictive accuracy across different CL strategies \cite{cossu2021}, but it was unknown for XAI techniques. Leveraging random recurrent models like ESNs effectively mitigates this effect.

We believe that continuously monitoring the explanations of a neural network is an important challenge that can impact both the design of XAI solutions and our understanding of what happens inside a CL model.

\begin{footnotesize}

\bibliographystyle{unsrt}
\bibliography{bib}

\end{footnotesize}

\end{document}